\def\year{2018}\relax
\documentclass[letterpaper]{article} %DO NOT CHANGE THIS
\usepackage{aaai18}  %Required
\usepackage{times}  %Required
\usepackage{helvet}  %Required
\usepackage{courier}  %Required
\usepackage{url}  %Required
\usepackage{graphicx}  %Required
\usepackage{amsmath}
\usepackage{multirow}
\usepackage{graphics}
\usepackage{bchart}
\usepackage{xcolor}
\usepackage{pgfplots}
\usepackage{tikz}
\usepackage{placeins}
\begin{document}

\title{Learning Graph-Level Representation for Drug Discovery}

\author{Junying Li, Deng Cai, Xiaofei He\\
State Key Laboratory of CAD\&CG, College of Computer Science and Technology \\
Zhejiang University\\
microljy@zju.edu.cn, dengcai@gmail.com, xiaofeihe@cad.zju.edu.cn\\
}

\maketitle
\begin{abstract}
Predicating macroscopic influences of drugs on human body, like efficacy and toxicity, is a central problem of small-molecule based drug discovery. Molecules can be represented as an undirected graph, and we can utilize graph convolution networks to predication molecular properties. However, graph convolutional networks and other graph neural networks all focus on learning node-level representation rather than graph-level representation. Previous works simply sum all feature vectors for all nodes in the graph to obtain the graph feature vector for drug predication. In this paper, we introduce a dummy super node that is connected with all nodes in the graph by a directed edge as the representation of the graph and modify the graph operation to help the dummy super node learn graph-level feature. Thus, we can handle graph-level classification and regression in the same way as node-level classification and regression. In addition, we apply focal loss to address class imbalance in drug datasets. The experiments on MoleculeNet show that our method can effectively improve the performance of molecular properties predication.
 
\end{abstract}
\section{Introduction}

%% social meaning
\noindent Reducing the high attrition rates in drug development continues to be a key challenge for the pharmaceutical industry. When biological researches yield evidence that a particular molecule could modulate essential pathways to achieve therapeutic activity, the discovered molecule often fails as a potential drug for a number of reasons including toxicity, low activity, and low solubility\cite{waring2015analysis}. The primary problem of small-molecule based drug discovery is finding analogue molecules with increased efficacy and reduced safe risks to the patient to optimize the candidate molecule. Machine learning is a powerful tool of drugs virtual screening that can help drug developer eliminate unqualified candidate molecules quickly.

%% now work
A molecule can be of arbitrary size and shape. However, most machine learning methods can only handle inputs of a fixed size. A wide-used conventional proposal is to utilize hand-crafted feature like ECFP\cite{rogers2010extended}, Coulomb Matrix\cite{rupp2012fast} as input and feed it into conventional classifier like random forest and multi-layer perceptron. In the last few years, convolutional neural network clearly surpassed the conventional methods that use hand-crafted feature and SVM in object classification tasks\cite{krizhevsky2012imagenet}. It indicates that end-to-end learning with reasonable differentiable feature extractor may surpass the conventional two stage classifier. Following this idea, \cite{duvenaud2015convolutional} model a molecule as a graph, the nodes of which stand for atoms and the edges of which represent chemical bonds linking some of the atoms together, and propose graph convolutional network which directly takes the graph as input to learn the representations of molecules. Previous studies show that graph convolutional network broadly offers the best performance in most of the datasets in MoleculeNet\cite{wu2017moleculenet}.

%% problem  
Molecular properties of drugs are almost macroscopic influences on human body, like efficacy and toxicity. In fact, unlike common graph-based application like web classification, citation network and knowledge graph, predicating drugs properties is graph-level classification and regression rather than node-level classification and regression. However, present graph neural networks, including graph convolutional network\cite{duvenaud2015convolutional}, similar graph networks that take the original molecule as input\cite{altae2017low,yao2017intrinsic,schutt2017quantum,gomes2017atomic}, and other general graph neural networks designed for all types of graph\cite{li2015gated,bruna2013spectral,defferrard2016convolutional,kipf2016semi,niepert2016learning}, all focus on learning node-level representation rather than graph-level representation. Previous works \cite{altae2017low,wu2017moleculenet} simply sums all feature vectors for all nodes in the graph to obtain the graph representation for molecule properties classification and regression. 

%% our work
To learn better graph-level representation and handle graph-level classification and regression, we introduce a dummy super node that is connected with all nodes in the graph by a directed edge, as the representation of the graph. Thus, we can handle graph-level classification and regression in the same way as node-level classification and regression. The idea of taking the feature of one specific node as the feature of the graph has been proposed before\cite{scarselli2009graph}, but earlier work simply assign the first atom in the atom-bond description as the specific node. In our work, the dummy super node is initialized to zero and get updated through graph convolutional networks simultaneously as the genuine nodes do. We modify the graph operations\cite{altae2017low} and apply node-level batch normalization to learn better features.
%Such modification could keep the genuine nodes learning local feature.
We take the feature of the dummy super node as the feature of the graph and feed it into classifier. 

Another problem of molecular properties predication is that the datasets of molecules are often unbalanced. That is, the positive samples may be only a small part of the total samples. Take HIV dataset\cite{hiv} as instance, there are only 1487 chemical compounds that are HIV active in the dataset of totally 41913 chemical compounds. Previous works\cite{altae2017low,wu2017moleculenet} did not pay much attention to the fact of unbalanced data. To address the unbalanced data, we adopt focal loss\cite{lin2017focal} and show that it can effectively improve the classification performance in unbalanced molecular dataset.

We evaluated our model on several datasets of toxity, biological activities and solubility in MoleculeNet, including Tox21 ToxCast, PCBA, MUV, HIV, FreeSolv. The results show that our method could effectively improve the performance on molecular properties predication.

%% result
\section{Related Work}
Our work draws inspiration both from the field of molecular machine learning and graph neural network. In what follows, we provide a brief overview of previous works in both fields.

\subsection{Molecule Mchine Learning}
Encoding molecules into fixed-length strings or vectors is a core challenge for molecular machine learning. \cite{rupp2012fast} introduce Coulomb Matrix as feature of molecules and apply a machine learning model to predict atomization energies of a diverse set of organic molecules. \cite{rogers2010extended} propose Extended-connectivity fingerprints (ECFP), a novel class of topological fingerprints for molecular characterization.

\cite{duvenaud2015convolutional} propose graph convolutional network that operates directly on graph, which could learn better representations than conventional method like ECFP. \cite{altae2017low} combine graph convolutional network with residual LSTM embedding for one-shot learning on drug discovery. \cite{yao2017intrinsic,schutt2017quantum} apply neural network to predicate the molecular energy and reduced the predication error to 1 kcal/mol. \cite{gomes2017atomic} propose atomic neural network to predicate the binding free energy of a subset of protein-ligand complexes found in the PDBBind dataset.\cite{ramsundar2015massively} applied Massively multitask neural architectures to synthesize information from many distinct biological sources. \cite{wu2017moleculenet} introduce MoleculeNet, a large scale benchmark for molecular machine learning, which contains multiple public datasets, and establish metrics for evaluation and high quality open-source implementations.

\subsection{Graph neural network}
Graph-based Neural networks have previously introduced in\cite{gori2005new,scarselli2009graph} as a form of recurrent neural network. \cite{li2015gated} modify the graph neural network by using gated recurrent units and modern optimization techniques and then extend to output sequences. Spectral graph convolutional neural networks are introduced by \cite{bruna2013spectral} and later extended by \cite{defferrard2016convolutional} with fast localized convolutions. \cite{kipf2016semi} introduced a number of simplifications to spectral graph convolutinal neural network and improve scalibility and classification performance in large-scale networks. \cite{cao2016deep} propose a novel model for learning graph representations, which generates a low-dimensional vector representation for each vertex by capturing the graph structural information. \cite{niepert2016learning} propose a framework for learning convolutional neural networks for arbitrary graphs and applied it to molecule classification. 

All above work focus on learning node-level representation in graph rather than graph-level representation. However, predicating molecular properties is in fact a problem of graph-level classification and regression. In this paper, we introduce a dummy super node that is connected to all other nodes by a directed edge to learn graph-level representation rather than simply using the sum of the vectors of all nodes as the graph-level representation\cite{duvenaud2015convolutional,altae2017low}.

\section{Learning Graph-Level Representation}
For most macroscopic molecular properties predication tasks, like efficacy and toxicity, we could neglect the detailed edge information and take a molecule as an undirected graph\cite{wu2017moleculenet}. We apply graph convolutional networks\cite{duvenaud2015convolutional} to learn representation of the atoms(nodes) in the molecule. To learn the representation of the molecule(graph), we introduce a dummy super node that is connected with all nodes in the graph by a directed edge. We modify the graph operation to help the dummy super node learn graph-level feature, and utilize neural network as a classifier for graph-level properties predication.

\subsection{Graph Convolutional Network}

\graphicspath{{img/}}
\begin{figure*}
\label{fig::graph}
\begin{center}
\includegraphics[width=0.8\textwidth]{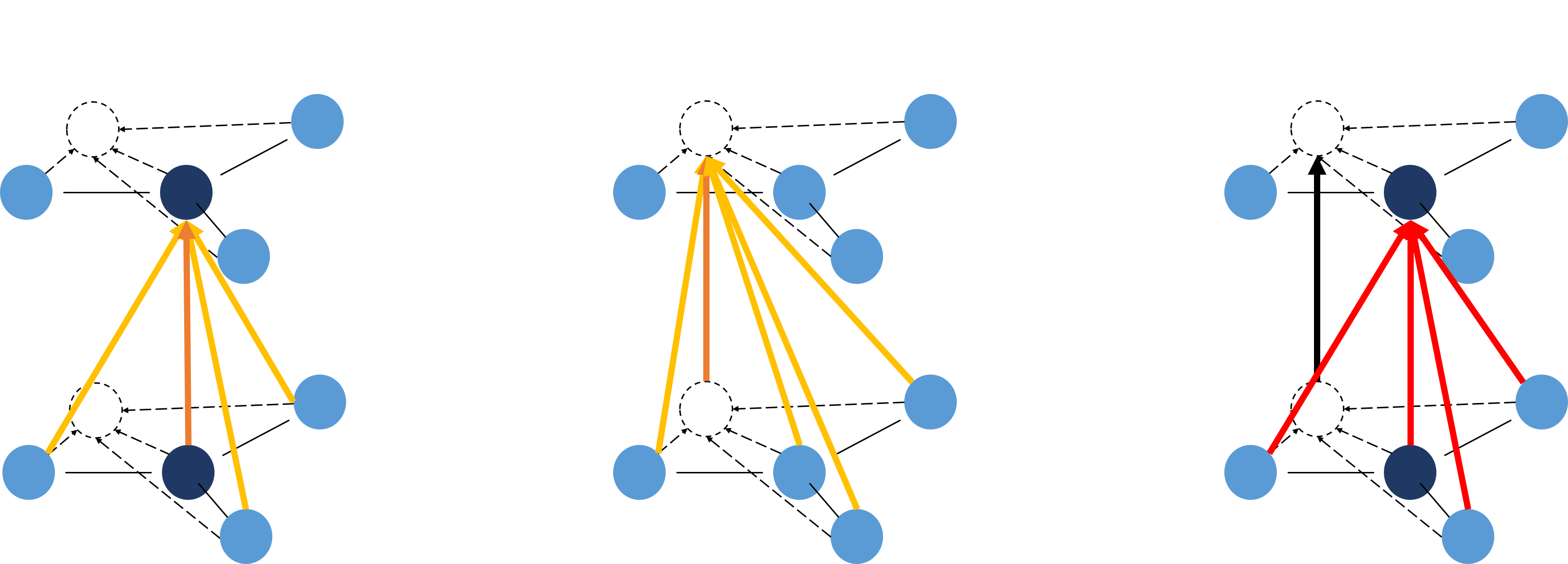}
\end{center}
\begin{picture}(0,0)
    \put(73,0){GraphConv}
    \put(205,0){GraphConv(Super Node)}
    \put(390,0){GraphPool}
    
    \put(90,28){$\boldsymbol{v}$}    
    \put(405,28){$\boldsymbol{v}$} 
    \put(237,60){$\boldsymbol{S}$} 
    \put(237,135){$\boldsymbol{S}$} 
    \put(396,60){$\boldsymbol{S}$} 
    \put(396,135){$\boldsymbol{S}$} 

    \put(56,160){\scriptsize{$\boldsymbol{v}$ and its neighbors are fed }}
    \put(54,150){\scriptsize{into dense layer and summed}}
    \put(217,160){\scriptsize{$\boldsymbol{S}$ and all nodes are fed }}
    \put(212,150){\scriptsize{into dense layer and summed}}
    \put(368,155){\scriptsize{Max over $\boldsymbol{v}$ and its neighbors}}
\end{picture}
\caption{\textbf{Graphical representation of the major graph operation.} The dark blue node represents the specific node $\boldsymbol{v}$, the light blue nodes represent the neighbours of $\boldsymbol{v}$, the dotted node represents the dummy super node $\boldsymbol{S}$. The orange and yellow arrows represent dense layer with different weights. The red arrows represent max polling, the black arrow represents identical mapping. The dummy super node is connected with all other nodes by a directed edge. The directed edge indicates that the dummy super node learns feature from genuine nodes but does not affect the feature of genuine nodes. Note that, for GraphConv and GraphPool, the operation is shown for a single node $\boldsymbol{v}$, however, these operations are performed on all nodes in graph simultaneously in fact.}
\end{figure*}
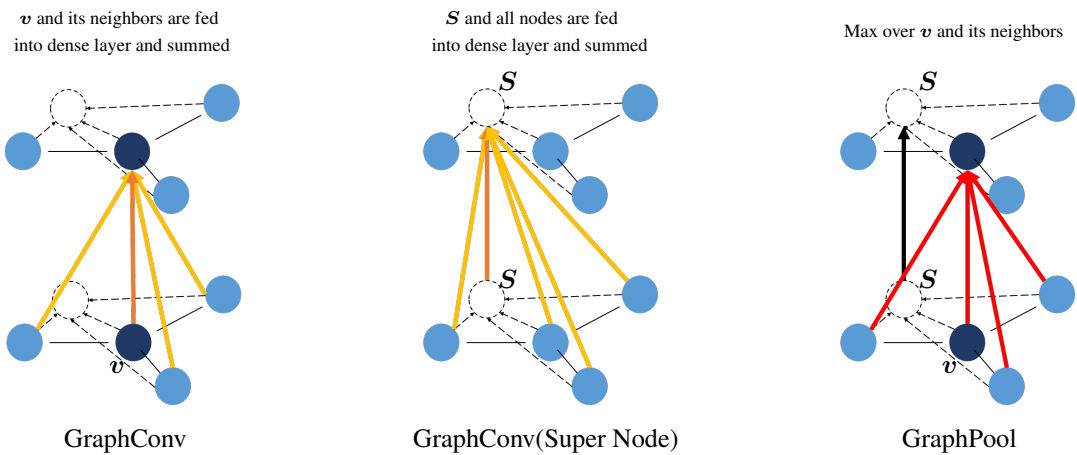

A molecule could be modeled as a graph, the nodes of which stand for atoms and the edges of which represent chemical bonds linking some of the atoms together. In graph convolution network, for a specific node, we separately feed it and its neighbours into two dense layer, and calculate the sum as the new features of the node. The weights are shared between dense layers that operate on different node with the same degree. Formally, for a specific node $\boldsymbol{v}$, that have totally $d$ neighbours $\boldsymbol{n_i}(1 \leq i \leq d)$, the new feature of the node $\boldsymbol{v'}$ of graph convolution operation is formulated as
\begin{equation}
\label{equ:node}
    \boldsymbol{v'} = \boldsymbol{W_{self}^d v} + \sum_{1 \leq i \leq d}\boldsymbol{W_{nb}^d n_i} + \boldsymbol{b_d}
\end{equation}

where $\boldsymbol{W_{self}^d}$ is the weight for self node, $\boldsymbol{W_{nb}^d}$ is the weight for neighbour node, $boldsymbol{b_d}$ is the bias. These weights vary from different degree $d$ of the specific node, for that the nodes with different number of neighbours are affected by neighbours in different ways. Since the output of graph convolution layer is also a graph, we could continually apply graph convolution layers to the graph. Thus, the receptive field of the network would be larger, and the nodes would be affected by further neighbours. The left thumbnail in figure \ref{fig::graph} illustrates the graph convolution operation. The yellow arrows indicate the dense layer with weight $\boldsymbol{W_{nb}^d}$ for neighbours, representing the neighbours' effect on the specific node. The orange arrows indicate the dense layer with weight $\boldsymbol{W_{self}^d}$ for the specific node self.

In analogy to pooling layers in convolutional neural networks, Graph Pooling is introduced by \cite{altae2017low}. Graph Polling is a simple operation that returns the maximum activation across the node and its neighbours. Simple illustration is shown in Figure \ref{fig::graph}. With graph pooling, we could easily enlarge the receptive field without adding extra weights. However, graph pooling does not change the size of the graph as the pooling of CNN does to the features of images.

Although there is no real convolution operation in graph convolutional network, graph convolutional network is design for similar reason to convolutional neural network, the 'local geometry' and reasonable weight sharing. Convolutional networks\cite{lecun1998gradient} apply convolutions to exploit the translation equivariance of image features. And the feature learned form CNN is translation equivariant to some extent. Graph convolutional network also focus on learning local feature, through sharing weight with dense layers operate on nodes with the same number of degree. Thus, the feature learned by graph convolutional is determined by local neighbours rather the global position of nodes. Such features are equivariant to permutation of atomic group.

\begin{figure*}
\label{fig::model}
\begin{center}
\includegraphics[width=0.94\textwidth]{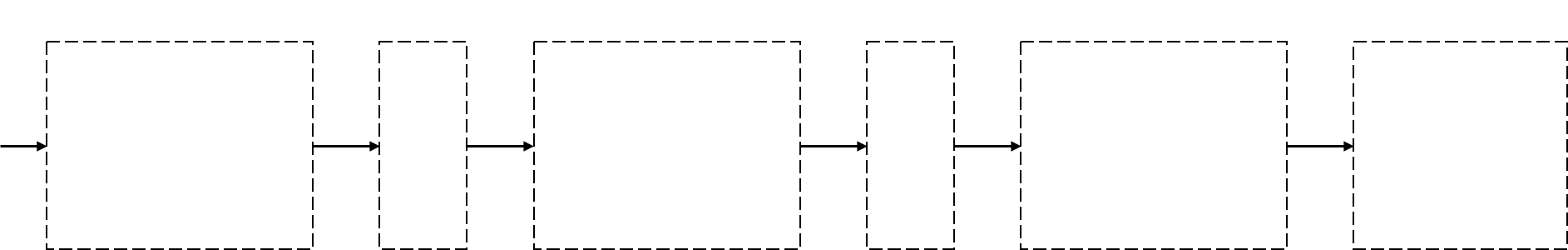}
\end{center}
\begin{picture}(0,0)
    \put(75,0){GC: Graph Convolution}
    \put(186,0){GP: Graph Pooling}
    \put(280,0){BN: Node-Level Batch Normalization}
    
    \put(37,44){GC+RELU+BN}
    \put(137,44){GP}
    
    \put(185,44){GC+RELU+BN}
    \put(285,44){GP}
    
    \put(332,44){GC+RELU+BN}
    \put(438,50){Two-Layer}
    \put(438,38){Classifier}
    
\end{picture}
\caption{\textbf{Our Network Structure.} We apply three graph convolution blocks(GC + RELU + BN) and two graph pooling layers, then we feed the feature of the dummy super node to a two-layer classifier. The batch normalization used here is node-level batch normalization. The third graph convolution block only calculates the feature of the dummy super node, for the new feature of other nodes would not be fed into following classifier.}
\end{figure*}
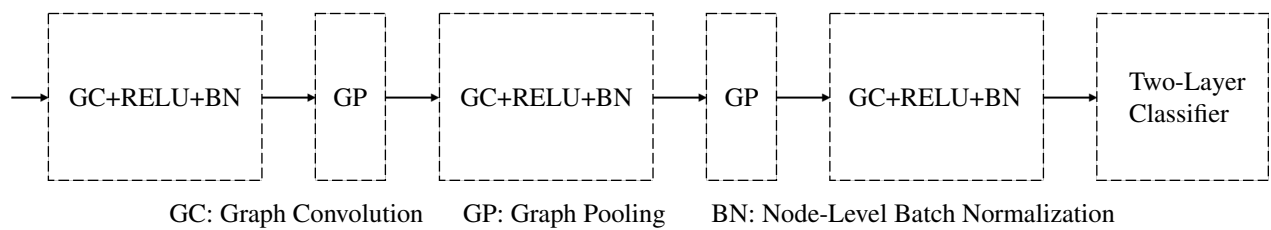

\subsection{Dummy Super Node}
Learning graph-level is a central problem of molecule classification and regression. Modern graph convolution networks often contain two graph convolution layers and two graph polling layers, the receptive field of the such model is too small, especially compared with the size of most drug compounds. Simply summing the node-level features can not enlarge the receptive of the model in a real sense. A naive way to enlarge the receptive field is simply applying more graph convolutional layers. However, such method makes model more likely to overfit and perform worse in validation and test subset for the lack of enough data in drug datasets.

In order to learn better graph-level representations, we introduce a dummy super node that is connected with all nodes in the graph by a directed edge, as the representation of the graph. Since the dummy super node is directly connected with all nodes in graph, it can easily learn global feature through one graph convolution layer. The directed edge pointed to the dummy super node from other genuine nodes, indicates that the dummy super node could learn features from all other genuine nodes, while none of the genuine nodes would be affected by the dummy super node. Consequently, the dummy super node could learn the global feature while the genuine nodes keep learning local features that is equivariant to permutation of atomic group. Since the feature of molecules is likely to be more complex than that of atoms, we could use a longer vector as the feature of the dummy super node. 

Formally, we could formulate the new feature of the dummy super node $\boldsymbol{S'}$ of graph convolution layer as 
\begin{equation}
\label{equ:dummy}
    \boldsymbol{S'} = \boldsymbol{W_{self} S} + \sum_{1 \leq i \leq N}\boldsymbol{W_{nb} v_i} + \boldsymbol{b}
\end{equation}
where $\boldsymbol{W_{self}}$ is the weight for the dummy super node, $\boldsymbol{W_{nb}}$ is the weight for genuine node, $N$ is the number of nodes, $\boldsymbol{v_i}$ is the feature of $i$th node, $\boldsymbol{b}$ is the bias. The dummy super node $\boldsymbol{S}$ is initialized to zero and get updated through graph convolutional networks simultaneously as the genuine nodes do.

The new features of genuine nodes are not affected by the dummy super node, and keep the same as Equation \ref{equ:node}. Graph pooling is only applied to genuine nodes, and the dummy super node is not regarded as neighbours of other nodes when graph pooling is applied.% Above modification is designed to keep genuine nodes learning node-level representation

%Then, we can handle graph-level classification in the same manner as node-level classification. 

%We modifies the graph operations\cite{altae2017low} to prevent the genuine nodes in graph affected by the dummy super node, while the dummy super node keep learning features from all genuine nodes. 
%Such modification could keep the genuine nodes learning local feature.
%We take the feature of the dummy super node as the feature of the graph and feed it into dense layer. 

\subsection{Network Structure}

In addition to graph operations, we also apply node-level batch normalization, relu activation, tanh activation and dense layer in the model. For some datasets with imbalance data, we replace common cross entropy loss with focal loss. The whole network structure is shown in Figure \ref{fig::model}. We add the dummy super node and its edges to the original graph before sending it to the network. In our model, We apply three graph convolution blocks(GC + RELU + BN) and two graph pooling layers, then we feed the feature of the dummy super node to a two-layer classifier. The batch normalization used there is node-level batch normalization that would be discussed later. Only  the feature of the dummy super node is calculated in the third graph convolution block, for the new feature of other nodes would not be fed into following classifier.

Unlike the ResNet, we place relu activation before batch normalization, for the experiment shows that it is hard to train when we place relu activation behind. We argue that the molecule feature is different from that of images. The noise in images usually does not affect the image label, while that in molecules not. 

\subsubsection{Node-Level Batch Normalization}
Since the number of atoms vary from molecules, standard batch normalization\cite{ioffe2015batch} can not be applied to graph convolution network directly. We extend standard batch normalization to node-level batch normalization. We normalize the feature of each node, make it zero mean and unit variance. Since the dummy super node is also a node, we simply apply identical normalization to the dummy super node as well.

\subsubsection{Focal Loss}
Data in drug datasets is often unbalanced. Take HIV dataset as instance, there are only 1487 chemical compounds that are HIV active in the dataset of totally 41913 chemical compounds. However, previous work\cite{altae2017low,wu2017moleculenet} do not pay much attention to the fact of unbalanced data in Drug dataset.

Downsample is a simple method to tackle with unbalanced data. However, the number of data in drug datasets is always not large for the difficulty of acquiring data and get them labeled, and simple downsample may result in serious overfitting of neural network for the lack of sufficient data. In addition, Boosting is an effective resample method that may address imbalance data. But it's not easy to integrate boosting with deepchem\cite{wu2017moleculenet}, the modern chemical machine learning framework. 

Focal Loss is an elegant and brief proposal that introduced by \cite{lin2017focal} to address class imbalance. It shows that focal loss could effectively improve the performance of one-stage detector in object detection. In fact, Focal Loss is a kind of reshaped cross entropy loss that the weighs of well-classified examples are reduced. Formally, Focal Loss is defined as 

$$FL(p)=-(y(1-p)^\gamma \log p+(1-y)p^\gamma \log (1-p)).$$

where $y \in \{0,1\}$ specifies the ground-truth class, $p \in [0,1]$ is the model's estimated probability for the class with label $y=1$, $\gamma$ is tunable focusing parameter. When $\gamma = 0$, focal loss is equivalent to cross entropy loss, and as $\gamma$ is increased the effect of the modulating factor is likewise increased. Focal loss focuses on training on a sparse set of hard examples and prevents the vast number of easy negatives from overwhelming the classifier during training.

%We argue that focal loss can boost not only the performance of unbalanced data classification, but also the performance of multi-label learning. A common problem encounted in multi-label learning is that most of classes are well-classified while other few are not. However, the gradient of most of classes that are well-classified could overwhelming that of the few number of classes that are not well-classified. Focal loss could improve the performance of multi-label learning by reducing the weight of well-classified classes.

\section{Experiment}
We evaluate our model on several datasets of MoleculeNet, ranging from drug activity, toxity, solvation. Our code is released in Github\footnote{\url{https://github.com/microljy/graph_level_drug_discovery}}.

\begin{table*}[!h]
\caption{\textbf{The Performance in FreeSolv dataset.} $R^2$ is squared Pearson correlation coefficient(larger is better). Since $R^2$ used by \cite{wu2017moleculenet} can not perfectly reflect the performance of the model, we provide with RMSE and MAE(kcal/mol) as well. }
\label{tbl::freesolv}
\vskip 0.15in
\begin{center}
\begin{small}
\begin{tabular}{|c|c|c|c|c|c|c|c|}
\hline
\multirow{2}{*}{Model}         & \multirow{2}{*}{Split Method}&  \multicolumn{3}{c|}{Valid}  &\multicolumn{3}{c|}{Test} \\
\cline{3-8}
                               &             & $R^2$ & RMSE   & MAE   & $R^2$ & RMSE  & MAE\\
\hline
\multirow{3}{*}{GraphConv}     & Index       & 0.935          & 0.909          & 0.703          & 0.941          & 0.963          & 0.738 \\
                               & Random      & 0.928          & \textbf{0.982}          & \textbf{0.644} & 0.895          & 1.228          & 0.803 \\
                               & Scaffold    & 0.883          & 2.115          & 1.555          & 0.709          & 2.067          & 1.535 \\
\hline
\multirow{3}{*}{\textbf{Our}}  & Index       & \textbf{0.952} & \textbf{0.787} & \textbf{0.566} & \textbf{0.945} & \textbf{0.933} & \textbf{0.598} \\
                               & Random      & \textbf{0.933} & 1.010          & 0.652          & \textbf{0.910} & \textbf{1.112} & \textbf{0.659} \\
                               & Scaffold    & \textbf{0.884} & \textbf{2.076} & \textbf{1.404} & \textbf{0.746} & \textbf{1.939} & \textbf{1.415} \\  
\hline
\end{tabular}
\end{small}
\end{center}
\vskip -0.1in
\end{table*}

\subsection{MoleculeNet} 
MoleculeNet is a dataset collection built upon multiple public databases, covering various levels of molecular properties, ranging from atomic-level properties to macroscopic influences on human body.

For chemical data, random splitting datasets into train/valid/test collections that is widely used in machine learning, is often not correct\cite{sheridan2013time}. Consequently, MoleculeNet implements multiple different splittings(Index Split, Random Split, Scaffold Split) for each dataset. Scaffold split attempts to separate structurally molecules in the training/validation/test sets, thus the scaffold split offers a greater challenge and demands a higher level of generalization ability for learning algorithms than index split and random split. 

For quite a few datasets in MoleculeNet, the number of the positive samples and the number of negative samples is not balanced. Thus the accuracy metrics widely used in machine learning classification tasks are not suitable here. In MoleculeNet, classification tasks are evaluated by area under curve (AUC) of the receiver operating characteristic (ROC) curve, and regression tasks are evaluated by squared Pearson correlation coefficient (R2).

We pick following datasets(Tox21, ToxCast, HIV, MUV, PCBA, FreeSolv) on macroscopic chemical influences on human body from MoleculeNet, and evaluate our model.

\subsubsection{Tox21} 
The “Toxicology in the 21st Century” (Tox21) initiative created a public database measuring toxicity of compounds, which has been used in the 2014 Tox21 Data Challenge\cite{tox21}. Tox21 contains qualitative toxicity measurements for 8014 compounds on 12 different targets, including stress response pathways and nuclear receptors.

\subsubsection{ToxCast}ToxCast is another data collection (from the same initiative as Tox21) providing toxicology data for a large library of compounds based on in virtual high-throughput screening\cite{richard2016toxcast}. The processed collection in MoleculeNet contains qualitative results of over 600 experiments on 8615 compounds.

\subsubsection{MUV}The MUV dataset\cite{rohrer2009maximum} contains 17 challenging tasks for around 90,000 compounds and is specifically designed for virtual screening techniques. The positives examples in these datasets are selected to be structurally distinct from one another.% As a result, this collection is a best-case scenario for baseline machine learning (since each data point is maximally informative).

\subsubsection{PCBA}
PubChem BioAssay (PCBA) is a database that consists of biological activities of small molecules generated by high-throughput screening\cite{wang2011pubchem}. The processed collection in MoleculeNet is a subset that contains 128 bioassays measured over 400,000 compounds.

\subsubsection{HIV}The HIV dataset was introduced by the Drug Therapeutics Program (DTP) AIDS Antiviral Screen, which tested the ability to inhibit HIV replication for 41,913 compounds\cite{hiv}. Screening results were evaluated and placed into three categories: confirmed inactive (CI), confirmed active (CA) and confirmed moderately active (CM). In MoleculeNet, the latter two labels are combined, making it a classification task between inactive (CI) and active (CA and CM).

\subsubsection{FreeSolv}The Free Solvation Database (FreeSolv) provides experimental and calculated hydration free energy of small molecules in water\cite{Mobley2014FreeSolv}. FreeSolv contains 643 compounds, a subset of which are also used in the SAMPL blind prediction challenge\cite{Mobley2014Blind}.

\subsection{Settings and Performance Comparisons}
We apply the identical network to all above datasets, except for the output channel number of the last layer that is determined by the number of tasks. For different datasets, we only tune the hyperparameter of training. We compare our model with logistic regression with ECFP feature as input and standard graph convolution network. To have a fair comparison, we reimplement graph convolution network, and most of the performance of our implements are much better than that reported by \cite{wu2017moleculenet}.

\subsubsection{Classification Tasks} For classification tasks(Tox21, ToxCast, MUV, PCBA), we utilize the area under curve (AUC) of the ROC curve as metric(larger is better). The results are shown in Table \ref{tbl::tox21}, Table \ref{tbl::toxcast}, Table \ref{tbl::muv} and Table \ref{tbl::pcba}. 
Our model generally surpasses standard graph convolution network and conventional logistic regression in both small datasets on toxity(Tox21, ToxCast) and larger datasets on bioactivities(MUV, PCBA). It indicates that our method could generally improve the performance of graph convolution network. Compared with standard graph convolution network, in average, our model achieves an improvement of about 1.5\% in the test subset of the four datasets.

\begin{table}[!h]
\caption{\textbf{The area under curve (AUC) of the ROC curve of various models in Tox21 dataset.} }
\label{tbl::tox21}
\vskip 0.15in
\begin{center}
\begin{small}
\begin{tabular}{|c|c|c|c|c|}
\hline
Model                           & Split Method    & Train         &Valid         & Test \\
\hline
\multirow{3}{*}{ECFP+LR}        & Index      &   0.903       &   0.704      &   0.738  \\
                                & Random     &   0.901       &   0.742      &   0.755  \\
                                & Scaffold   &   0.905       &   0.651      &   0.697  \\
\hline
\multirow{3}{*}{GraphConv}      & Index      &   0.945       &   0.829      &   0.820  \\
                                & Random     &   0.938       &   0.833      &   0.846  \\
                                & Scaffold   &   0.955       &   0.778      &   0.752  \\
\hline
\multirow{3}{*}{\textbf{Our}}   & Index      &\textbf{0.965} &\textbf{0.839} &\textbf{0.848}\\
                                & Random     &\textbf{0.964} &\textbf{0.842} &\textbf{0.854}\\
                                & Scaffold   &\textbf{0.971} &\textbf{0.788} &\textbf{0.759}\\  
\hline
\end{tabular}
\end{small}
\end{center}
\vskip -0.1in
\end{table}

\begin{table}[!h]
\caption{\textbf{The area under curve (AUC) of the ROC curve of various models in ToxCast dataset.} }
\label{tbl::toxcast}
\vskip 0.15in
\begin{center}
\begin{small}
\begin{tabular}{|c|c|c|c|c|}
\hline
Model                           & Split Method    & Train         &Valid         & Test \\
\hline
\multirow{3}{*}{ECFP+LR}        & Index      &   0.727       &   0.578      &   0.464  \\
                                & Random     &   0.713       &   0.538      &   0.557  \\
                                & Scaffold   &   0.717       &   0.496      &   0.496  \\
\hline
\multirow{3}{*}{GraphConv}      & Index      &   0.904       &   0.723      &   0.708  \\
                                & Random     &   0.901       &   0.734      &   0.754  \\
                                & Scaffold   &   0.914       &   0.662      &   0.640  \\
\hline
\multirow{3}{*}{\textbf{Our}}   & Index      &\textbf{0.927} &\textbf{0.747} &\textbf{0.734}\\
                                & Random     &\textbf{0.924} &\textbf{0.746} &\textbf{0.768}\\
                                & Scaffold   &\textbf{0.929} &\textbf{0.696} &\textbf{0.657}\\  
\hline
\end{tabular}
\end{small}
\end{center}
\vskip -0.1in
\end{table}

\begin{table}[!h]
\caption{\textbf{The area under curve (AUC) of the ROC curve of various models in PCBA dataset.} }
\label{tbl::pcba}
\vskip 0.15in
\begin{center}
\begin{small}
\begin{tabular}{|c|c|c|c|c|}
\hline
Model                           & Split Method    & Train         &Valid         & Test \\
\hline
\multirow{3}{*}{ECFP+LR}        & Index      &   0.809       &   0.776      &   0.781  \\
                                & Random     &   0.808       &   0.772      &   0.773  \\
                                & Scaffold   &   0.811       &   0.746      &   0.757  \\
\hline
\multirow{3}{*}{GraphConv}      & Index      &   0.895       &   0.855      &   0.851  \\
                                & Random     &   0.896       &   0.854      &   0.855  \\
                                & Scaffold   &   0.900       &   0.829      &   0.829  \\
\hline
\multirow{3}{*}{\textbf{Our}}   & Index      &\textbf{0.904} &\textbf{0.869} &\textbf{0.864}\\
                                & Random     &\textbf{0.899} &\textbf{0.863} &\textbf{0.867}\\
                                & Scaffold   &\textbf{0.907} &\textbf{0.847} &\textbf{0.845}\\  
\hline
\end{tabular}
\end{small}
\end{center}
\vskip -0.1in
\end{table}

\begin{table}[!h]
\caption{\textbf{The area under curve (AUC) of the ROC curve of various models in MUV dataset.} }
\label{tbl::muv}
\vskip 0.15in
\begin{center}
\begin{small}
\begin{tabular}{|c|c|c|c|c|}
\hline
Model                           & Split Method    & Train         &Valid         & Test \\
\hline
\multirow{3}{*}{ECFP+LR}        & Index      &   0.960       &   0.773      &   0.717  \\
                                & Random     &   0.954       &   0.780      &   0.740  \\
                                & Scaffold   &   0.956       &   0.702      &   0.712  \\
\hline
\multirow{3}{*}{GraphConv}      & Index      &   0.951       &\textbf{0.816}&   0.792  \\
                                & Random     &   0.949       &   0.787      &   0.836  \\
                                & Scaffold   &   0.979       &   0.779      &   0.735  \\
\hline
\multirow{3}{*}{\textbf{Our}}   & Index      &\textbf{0.982} &\textbf{0.816} &\textbf{0.795}\\
                                & Random     &\textbf{0.989} &\textbf{0.800} &\textbf{0.845}\\
                                & Scaffold   &\textbf{0.990} &\textbf{0.816} &\textbf{0.762}\\  
\hline
\end{tabular}
\end{small}
\end{center}
\vskip -0.1in
\end{table}

\begin{table}[!h]
\caption{\textbf{The area under curve (AUC) of the ROC curve of various models in HIV dataset.} With focal loss($\gamma = 2$), our method achieves further improvement.}
\label{tbl::hiv}
\vskip 0.15in
\begin{center}
\begin{small}
\begin{tabular}{|c|c|c|c|c|}
\hline
Model                               & Split Method    & Train         &Valid         & Test \\
\hline
\multirow{3}{*}{ECFP+LR}           & Index      &   0.864       &   0.739      &   0.741  \\
                                   & Random     &   0.860       &   0.806      &   0.809  \\
                                   & Scaffold   &   0.858       &   0.798      &   0.738  \\
\hline
\multirow{3}{*}{GraphConv}         & Index      &   0.945       &   0.779      &   0.728  \\
                                   & Random     &   0.939       &   0.835      &   0.822  \\
                                   & Scaffold   &   0.938       &   0.795      &   0.769  \\
\hline
\multirow{3}{*}{Our}               & Index      &   0.973       &   0.789       &   0.737\\
                                   & Random     &   0.951       &   0.842       &   0.830\\
                                   & Scaffold   &   0.967       &   0.813       &   0.763       \\  
\hline
\multirow{3}{*}{\textbf{Our + Focal Loss}}  & Index      &\textbf{0.993} &\textbf{0.793} &\textbf{0.749}\\
                                   & Random     &\textbf{0.993} &\textbf{0.843} &\textbf{0.851}\\
                                   & Scaffold   &\textbf{0.992} &\textbf{0.816} &\textbf{0.776}\\                                    
\hline
\end{tabular}
\end{small}
\end{center}
\vskip -0.1in
\end{table}

\subsubsection{Regression Tasks} For regression tasks(FreeSolv), we utilize squared Pearson correlation coefficient($R^2$), RMSE, MAE as metrics. The experimental results are shown in Table \ref{tbl::freesolv}. We only report the performance of graph convolution network and our model here. Our model has a better performance in general. We argue that our model failed to surpass graph convolution network in validation subset under random splitting for stochastic factors introduced by splitting method and too few numbers of compounds in FreeSolv.

It is interesting that how our method competes with classic ab-initio calculations. Hydration free energy that should be predicated in FreeSolv has been widely used as a test of computational chemistry methods. With energy values ranging from -25.5 to 3.4kcal/mol in the FreeSolv dataset, the RMSE of ab-initio calculations results reach up to 1.5kcal/mol\cite{Mobley2014Blind}. Our methods clearly outperform ab-initio calculations when we utilize index split and random split. However, our method fails to outperform ab-initio calculations when the dataset is split by scaffold split that separate molecules structurally. We argue that it is the insufficient data that results in weak generalization ability of our model. When fed with enough data, our model may overall surpass classic ab-initio calculations.

\subsubsection{imbalance Class} We apply focal loss in HIV dataset that have unbalanced class. The experimental results of Table \ref{tbl::hiv} shown that focal loss could further improve the performance. It is interesting that our model overfits quickly in train dataset when we apply focal loss. We argue that focal loss helps our model handle hard example and fit better in train dataset.

\iffalse

\definecolor{bblue}{HTML}{4F81BD}
\definecolor{rred}{HTML}{C0504D}
\definecolor{ggreen}{HTML}{9BBB59}
\definecolor{ppurple}{HTML}{9F4C7C}

\begin{tikzpicture}
    \begin{axis}[
        width  = 0.5*\textwidth,
        height = 14cm,
        major y tick style = transparent,
        xbar,
        bar width=5pt,
        xmajorgrids = true,
        xlabel = {AUC},
        symbolic y coords={a,b,c,d,e,f,g,h,i,j,k,l,m,n,o,p,q},
        ytick = data,
        scaled x ticks = false,
    ]
        \addplot[style={bblue,fill=bblue,mark=none}]
            coordinates {(0.657,a) (0.972,b) (0.974,c) (0.854,d) (0.534,e) (0.878,f) (0.431,g) (0.869,h) (0.704,i) (0.806,j) (0.756,k) (0.831,l) (0.908,m) (1.0,n) (0.970,o) (0.670,p) (0.705,q) };

        \addplot[style={rred,fill=rred,mark=none}]
             coordinates {(0.512,a) (0.995,b) (0.908,c) (0.929,d) (0.704,e) (0.935,f) (0.271,g) (0.870,h) (0.883,i) (0.866,j) (0.756,k) (0.929,l) (0.955,m) (1.0,n) (0.982,o) (0.631,p) (0.837,q)};

        \legend{Baseline, Focal Loss}
    \end{axis}
\end{tikzpicture}
\fi

\FloatBarrier
\section{Conclusion}
In this paper, we point out that molecular properties predication demand graph-level representation. However, most of the previous works on graph neural network only focus on learning node-level representation. Such representation is clearly not sufficient for molecular properties predication. In order to learn graph-level representation, we propose the dummy super node that is connected with all nodes in the graph by a directed edge and modify the graph operation to help the dummy super node learn graph-level feature. Thus, we can handle graph-level classification and regression in the same way as node-level classification and regression. The experiment on MoleculeNet shows that our method could generally improve the performance of graph convolution network. 
\bibliography{graph-level.bbl}
\bibliographystyle{aaai}
%\printbibliography
\end{document}